# Improving Commonsense Bias Classification by Mitigating the Influence of Demographic Terms

**JinKyu Lee [1], Jihie Kim [1]**

[1]Department of Artificial Intelligence, Dongguk University, Seoul, Korea

Corresponding author: Jihie Kim(jihie.kim@dgu.edu).

This research was supported by the MSIT(Ministry of Science and ICT), Korea, under the ITRC(Information Technology Research Center) support program(IITP-2024-2020-0-01789), and the Artificial Intelligence Convergence Innovation Human Resources Development (IITP-2024-RS-2023-00254592) supervised by the IITP(Institute for Information & Communications Technology Planning & Evaluation).

**ABSTRACT** Understanding commonsense knowledge is crucial in the field of Natural Language Processing (NLP). However, the presence of demographic terms in commonsense knowledge poses a potential risk of compromising the performance of NLP models. This study aims to investigate and propose methods for enhancing the performance and effectiveness of a commonsense polarization classifier by mitigating the influence of demographic terms. Three methods are introduced in this paper : (1) hierarchical generalization of demographic terms (2) threshold-based augmentation and (3) integration of hierarchical generalization and threshold-based augmentation methods(IHTA). The first method involves replacing demographic terms with more general ones based on a term hierarchy ontology, aiming to mitigate the influence of specific terms. To address the limited bias-related information, the second method measures the polarization of demographic terms by comparing the changes in the model's predictions when these terms are masked versus unmasked. This method augments commonsense sentences containing terms with high polarization values by replacing their predicates with synonyms generated by ChatGPT. The third method combines the two approaches, starting with threshold-based augmentation followed by hierarchical generalization. The experiments show that the first method increases the accuracy over the baseline by 2.33%, and the second one by 0.96% over standard augmentation methods. The IHTA techniques yielded an 8.82% and 9.96% higher accuracy than threshold-based and standard augmentation methods, respectively.

## I. INTRODUCTION

Commonsense knowledge is a crucial part of human cognition. It encompasses a broad range of knowledge, such as our understanding of physicality, social dynamics and psychology, along with ability to anticipate and interpret human behavior [1]. As artificial intelligence systems continue to advance, the significance of commonsense knowledge in understanding natural language has become increasingly apparent.

The incorporation of commonsense knowledge into models has emerged as a growing trend, in various real-life scenarios [2]. Particularly notable within this trend is the recent emphasis on investigating biases associated with commonsense understanding [3].

Previous research highlighted the impact of demographic factors (such as gender, race, and ethnicity) on the performance of commonsense polarization classifiers [4,5]. Additionally, efforts to address unintended bias in NLP models have been made [6]. However, these methods have limitations in generalizability due to their focus primarily on specific classes of demographics, rather than comprehensively addressing broader spectrum of demographic influences. Futhermore, as shown in Fig. 1., language models may misclassify bias due to the influence of demographic terms. For example, the commonsense sentence "Iraq causes war" was classified by the BERT model as "Neutral" whereas humans labeled it as "Negative". On the other hand, as we just replace the country name, "Sweden causes war" was classified as "Negative" by BERT as well as humans. To enhance the ability of NLP systems to generate fair and unbiased outcomes, it is necessary to employ suitable methods that mitigate the influence of such demographic terms while preserving the integrity of the existing meaning and information.

We conducted an analysis of commonsense sentences, carefully considering the nuanced application of demographic terms. By examining the specific contexts in which these terms are used, we develop a method that improves their uses in commonsense polarization classifiers. This can facilitate more accurate and unbiased predictions by minimizing potential biases introduced by demographic terms.

The first method involves replacing the specific identity of demographic terms with more general terms, such as substituting 'Afghanistan' with 'Asian' and 'Asian' with 'people,' based on a predefined term hierarchy ontology. The second one measures the polarization of demographic terms by comparing the changes in the model's predictions when these terms are masked versus unmasked. Data augmentation is then applied to commonsense sentences with highly polarized terms by replacing their tuple relation with synonyms generated by ChatGPT. These methods demonstrate performance improvements over the baseline and standard data augmentation techniques, respectively. The third method combines the two approaches, starting with threshold-based augmentation followed by hierarchical generalization.

The contribution of this research is as follows. We introduce a novel approach that addresses demographic-based polarization bias in commonsense understanding. This approach entails analyzing language usage, considering the nuanced application of demographic terms, and designing methods to minimize their influence on commonsense polarization classifiers. Specifically, we propose two techniques: hierarchical generalization of demographic terms and threshold-based augmentation of commonsense sentences, and demonstrate their effectiveness.

## II. RELATED WORK

Extensive research in Natural Language Processing (NLP) has focused on investigating biases in language models and developing strategies to understand and mitigate them. A study exploring the relationship between the size of language models and their biases was conducted, finding that larger models exhibit more nuanced biases and an increased potential for bias in specific categories[5]. Researchers have also explored bias evaluation in language models, employing extrinsic evaluation methods and debiasing techniques to identify and address bias in real world applications and downstream tasks [7,8]. Intrinsic evaluation methods, such as analyzing word or sentence embeddings, contribute to a deeper understanding of biases within language models [9,10,11]. Additionally, It has been shown that systematic differences in bias measurement can be revealed through the unification of extrinsic evaluation methods, attributed to parameter choices [12]. Moreover, these methods have been used to identify biased decisions from models in specific tasks [13].

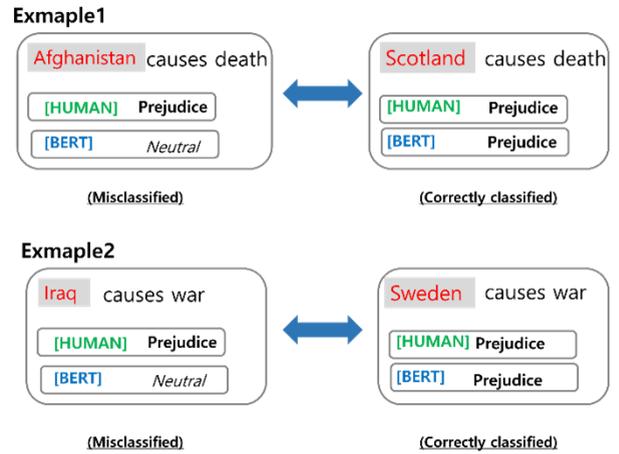

FIGURE 1. Example of Error in Commonsense Sentence Classification Resulting from the Influence of Demographic Terms.

Recent studies have drawn significant attention to the bias caused by demographic terms in language models. Various techniques have been proposed to address and mitigate demographic bias, aiming to promote fairness, equality, and inclusivity. These techniques include debiasing models trained on human-annotated examples[10], fairness-aware neural language models [14], bias-mitigating transformer architectures [15], and debiasing frameworks to identify and correct demographic bias [16]. Additionally, post-hoc debiasing techniques have been explored, where a debiasing step is added to sentence representations after initial training, prior to their utilization in downstream tasks [10,17]. While preceding investigations focused on biases in language models linked to particular categories like origin and gender, our study adopts a more extensive methodology, assessing biases across a wider range of demographic terms.

Building upon previous studies, our research proposes novel approaches to mitigate the impact of demographic terms and enhance the performance of the commonsense polarization classifier in knowledge models. This approach distinguishes itself from prior research by employing augmentation to substitute the predicate segment of a sentence with a synonym while retaining the original semantic content. Furthermore, to address bias stemming from demographic terms, our methodology involves substituting specific demographic terms with broader alternatives.

## III. Methodology

*A. Dataset*

A commonsense dataset, specifically chosen for its crowd-worker annotations, was utilized in this study [4]. Table 1. presents the categorization of the dataset into four groups: Gender, Origin, Profession, and Religion. This categorization provides an overview of the types of demographic terms included and their respective distributions in the dataset. To evaluate the polarization aspect from a human perspective, the dataset was crowd-labeled through Amazon Mechanical Turk. The unique dataset of 3,227 sentences within the dataset underwent evaluation by three workers, resulting in a total of 10,581 labeled sentences. The reliability of the crowd-sourced labels is indicated by Fleiss' kappa scores of 0.5007 and 0.3827, suggesting moderate agreement among annotators [18].

TABLE 1. Demographic terms statistics used in dataset

|  | Demograhpic terms | Method |
|---|---|---|
| Origin | American,African,Korea,etc. | 4474(42.28) |
| Religion | Muslim,Christian,etc. | 384(3.6) |
| Gender | Woman,son,brother,etc | 1565(14.80) |
| Profession | Doctor,Author,Athlete,etc. | 4158(39.29) |
| Total | - | 10581(100.00) |
| Unique | - | 3227 |

### B. Bias Classifier

The model is designed to perform polarization classification on the given dataset. It can be represented as $f \circ g : X \rightarrow R|Y|$, where $g(\cdot)$ is the feature extraction process using a language model like BERT, RoBERTa[19] and ChatGPT[20], $f(\cdot)$ represents the classification function, and Y represents the possible labels for the task. We classify commonsense sentences into three sentiment classes: Positive, Neutral, and Negative. To evaluate the model, we use standard metrics, including accuracy, recall, F-1 score, and precision. Fig. 2. shows the results of the bias classifier on demographic terms in the base dataset. In the "gender" category, accuracy, recall, F-1 score, and precision are 0.68, 0.68, 0.55, and 0.46 respectively. Similarly, for "profession", accuracy, recall, F-1 score, and precision are 0.73, 0.73, 0.66, and 0.64. "Origin" category has accuracy, recall, F-1 score, and precision of 0.87, 0.87, 0.85, and 0.87 respectively. Lastly, in "religion" category, accuracy, recall, F-1 score, and precision are 0.67, 0.67, 0.53, and 0.44 respectively. The classification metrics are relatively lower for categories with smaller dataset sizes, such as gender or religion, compared to larger categories like profession or origin.

### C. Debiasing Techniques

In this section, we introduce and detail various debiasing techniques aimed at reducing demographic biases in NLP models. These techniques are designed to identify, analyze, and mitigate the impacts of inherent biases present within language processing systems.

#### 1) HIERARCHICAL GENERALIZATION

We introduce a hierarchical generalization process aimed at mitigating the influence of demographic terms in commonsense bias classifiers while preserving the overall meaning and sentence structure. Our proposed approach builds upon the principles of Anonymization [18] and focuses on generalizing demographic terms within sentences while retaining their underlying semantic structure. The main goal of this technique is to alleviate the bias introduced by demographic terms in commonsense classifiers. The hierarchical structure consists of two levels: Hierarchy1 and Hierarchy2. Hierarchy1 is the first level of generalization, where specific demographic terms are replaced with more general terms. We adopted a categorization framework based on the criteria of the US labor force [21] to classify demographic terms into two hierarchical classes, as shown in Fig. 3. The constructed hierarchical structure encompassed categories related to religion, origin, profession, and gender. As we progressed through the levels of the hierarchy, the structure exhibited increasing levels of generalization. The utilization of a hierarchical classification approach enabled a more comprehensive analysis and understanding of similarities and differences across demographic groups. For example, terms such as 'she', 'lady', 'female', and 'stepmother' are replaced with the term 'woman'. Hierarchy2 is the second level of generalization, where Hierarchy1 is further generalized. Fig. 4. demonstrates the process of generalizing demographic terms based on the hierarchical structure. For instance, 'Afghanistan' is replaced with 'Asian', and 'Asian' is replaced with 'People', following the predefined hierarchy. Similarly, 'Korean' is replaced with 'Asian' and 'Asian' is replaced with 'People' according to the criteria of the US Labor Bureau data[21], which constitutes a predefined term hierarchy ontology. The hierarchical generalization method was applied to the entire dataset, the unique dataset, and the four categories within the unique dataset (Gender, Religion, Profession, Origin), replacing specific demographic terms with more general ones while maintaining content consistency.

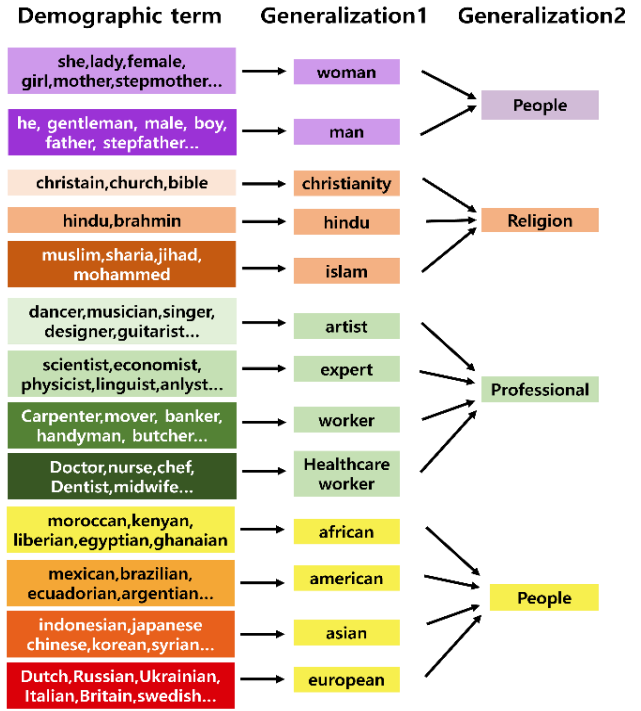

**FIGURE 2.** Generalization Process of Demographic Terms through Hierarchy Changes According to the Criteria of the US Labor Bureau Data.

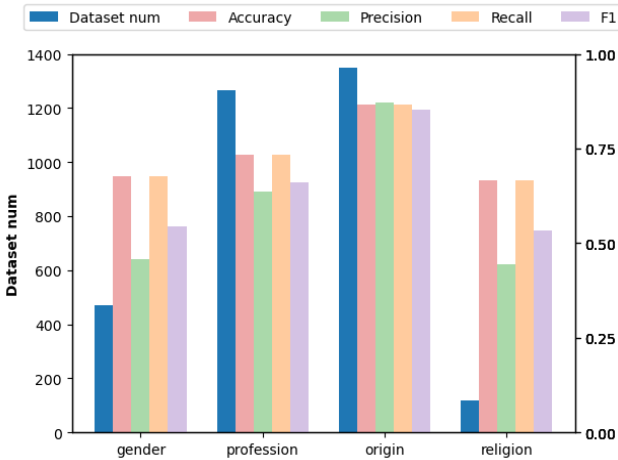

**FIGURE 3.** Performance of Bias Classifier on Demographic Term Categories in Base Dataset.

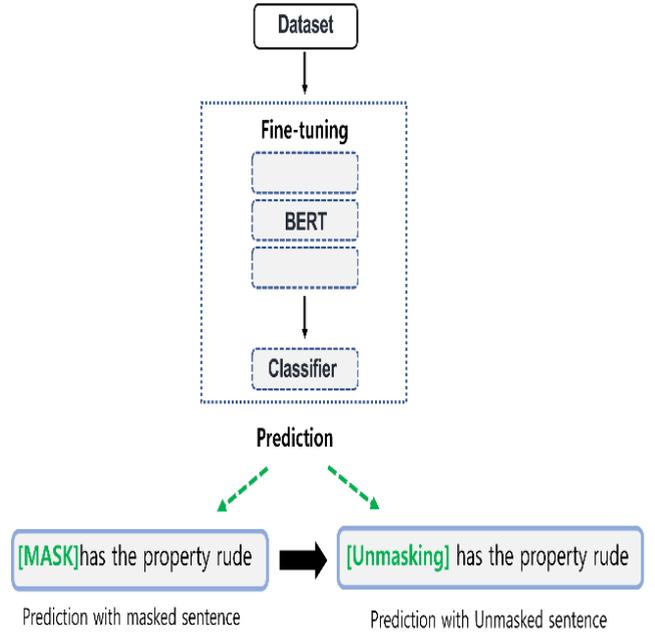

FIGURE 4. Process for Quantifying the Polarization of Demographic Terms in Commonsense Sentences

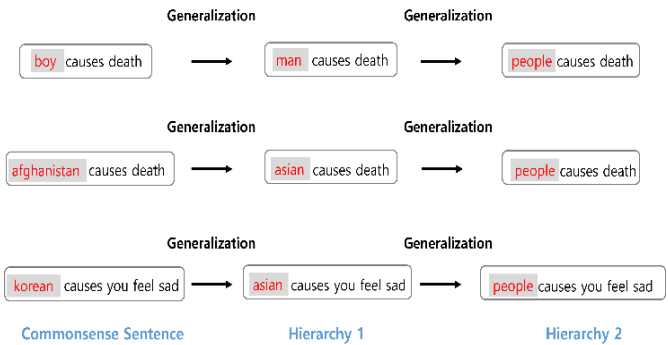

FIGURE 5. Hierarchical Generalization Process for Demographic Terms.

2) AUGMENTING COMMONSENSE SENTENCES BASED ON POLARIZATION OF DEMOGRAPHIC TERM

Previous research has predominantly focused on assessing sentiment associated with demographic terms to detect biases. However, there has been limited exploration of polarity measurement techniques that are used in other tasks. Our study leverages polarity measurement techniques in data augmentation in order to enhance the performance of commonsense classification. Data augmentation plays a crucial role in addressing limited bias datasets by increasing their size and addressing imbalanced data distribution, as shown in Fig. 2. We focus on augmenting samples based on polarized values identified through our threshold-based measurement. The proposed method measures the polarization of demographic terms by comparing the changes in the model's predictions. Our threshold-based measurement technique allows for the identification of sentiments, attitudes, and biases associated with demographic terms, thereby enhancing our understanding of the biases present in the data. To quantify the polarization of demographic terms in commonsense sentences, we follow the procedure shown in Fig. 5. We first finetune a BERT model on the commonsense dataset, and then predict commonsense sentences using the trained model, obtaining one polarity for the original sentences with unmasked demographic terms and the other polarity for the sentences after masking the demographic terms. We then compare the predicted polarities between the masked and unmasked instances of the same commonsense sentences to measure the polarity for a specific demographic term, as shown in Fig. 6. The polarization of a specific demographic term is determined by calculating the absolute differences between the number of sentences where its polarity changed in favor of the demographic term "N_changed_Positive" and the number of sentences where its polarity changed in negative against the demographic term "N_changed_Negative". This value is then divided by the total number of sentences containing the term "N_total". The resulting ratio, denoted as P, quantifies the polarization as follows:

$$P = |N\_changed\_Positive - N\_changed\_Negative| / N\_total \quad (1)$$

P indicates the polarity shift in the commonsense sentences where the term is included and can offer insights into the biases present in the data. In cases where the absolute value of P is greater than a threshold, it indicates that the specific demographic term can present more influence in determining the bias. In order to obtain more bias related sentences, a data augmentation is applied for demographic terms with higher P values.

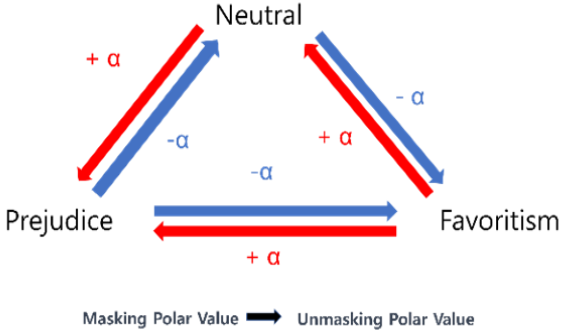

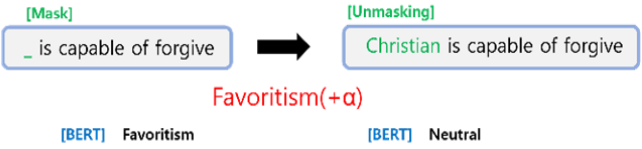

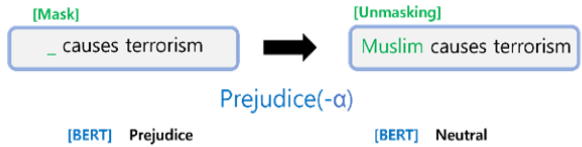

**FIGURE 7.** Polarity Analysis of demographic term in Commonsense Sentences: comparing the changes when Demographic term are masked vs. unmasked.

Generally, data augmentation involves enhancing the performance of classifiers by augmenting datasets that are overall limited or have insufficient representation for specific classes [22]. Therefore, in this research, we perform augmentation specifically targeting the Gender (14.80%) and Religion (3.6%) classes, which have relatively limited data, within a commonsense dataset composed of 4 demographic term classes: Gender, Religion, Profession, and Origin, as illustrated in Table 1. In particular, instead of uniformly replacing predicate portions for augmentation, we adopt a selective approach based on the influence of demographic terms on a polarization classifier. The augmentation is applied to commonsense sentences containing demographic terms with absolute polar values greater than 0, encompassing both positive and negative polarities, as demonstrated in Fig. 7.

The existing commonsense augmentation, such as counterfactual augmentation[8] and word pair utilization [23], aim to replace demographic terms. However, preserving the original meaning of the data presents a significant challenge. To address this challenge, we propose a threshold-based augmentation approach that maintains the original meaning and contextual information of commonsense by substituting the predicate instead of the subject or object during augmentation.

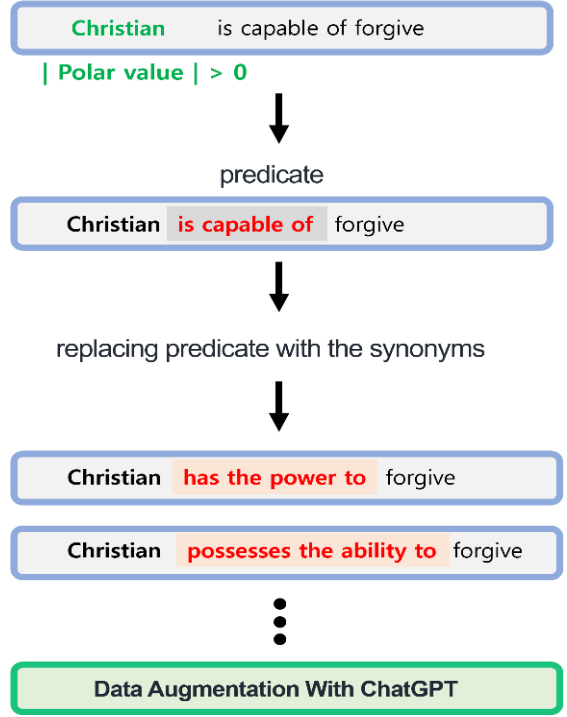

**FIGURE 6.** Threshold-based Commonsense Augmentation with Synonymous Predicate Replacement using ChatGPT.

This approach ensures dataset diversity and maintains semantic relationships, as illustrated in Fig. 7. To validate the effectiveness of our proposed technique, we conducted experiments using two augmentation methods: (i) threshold-based data augmentation and (ii) standard data augmentation. Data augmentation techniques often demonstrated their effectiveness in addressing class imbalances [22]. In the first method, we focused on augmenting the class with a relatively small number of instances. This involved selecting commonsense sentences that contained a demographic term surpassing the predefined threshold. We applied our proposed formulation, as described in Eq(1), to measure the influence of demographic terms. This strategy specifically targeted the augmentation of commonsense sentences containing demographic terms exceeding the threshold. By selectively augmenting commonsense sentences containing demographic terms with absolute polar values greater than 0, which exhibited a change in polarity after masking the demographic terms, we may enhance the performance of the polarization classifier. For the effective augmentation of commonsense sentences, we utilized ChatGPT to maintain the original meaning and context while introducing controlled variations. ChatGPT maintains semantic consistency and generates samples akin to human-labeled data. Consequently, employing ChatGPT for Synonym Augmentation enables more diverse and higher-quality data generation, surpassing not only traditional ontology-based methods but also backtranslation and word vector interpolation in latent space[24]. This approach aimed to preserve the original meaning and contextual information while introducing alternative expressions. The use of ChatGPT allowed efficient identification of synonym.

**TABLE 1.** Demographic terms statistics used in dataset

|  | Baseline | | | | Generalization1 | | | | Generalization2 | | | |
| --- | --- | --- | --- | --- | --- | --- | --- | --- | --- | --- | --- | --- |
|  | ACCURACY | PRECISION | RECALL | F-1 | Accuracy | Precision | Recall | F-1 | Accuracy | Precision | Recall | F-1 |
| Origin | 85.92 | 85.28 | 85.92 | 85.32 | 85.81 | 85.20 | 85.81 | 85.25 | 85.92 | 85.40 | 85.92 | 85.48 |
| Religion | 66.67 | 44.44 | 66.70 | 53.33 | 66.67 | 44.44 | 66.70 | 53.33 | 66.67 | 44.44 | 66.67 | 53.33 |
| Gender | 85.30 | 84.94 | 85.30 | 85.01 | 85.30 | 84.99 | 85.30 | 85.05 | 85.62 | 85.43 | 85.62 | 85.50 |
| Profession | 80.05 | 79.62 | 80.05 | 79.74 | 80.05 | 79.69 | 80.05 | 79.79 | 80.65 | 80.29 | 80.65 | 80.39 |
| Total | 82.43 | 81.96 | 82.43 | 81.76 | 82.62 | 82.07 | 82.62 | 82.17 | 87.4 | 88.01 | 87.40 | 87.58 |
| Unique | 84.50 | 86.73 | 84.05 | 85.00 | 86.36 | 87.29 | 86.36 | 87.0 | 87.4 | 88.01 | 87.40 | 88.00 |

#### 3) INTEGRATED HIERARCHICAL GENERALIZATION AND THRESHOLD-BASED AUGMENTATION (IHTA)

Previous research has predominantly focused

We propose a combined approach, termed Integrated Hierarchical and Threshold-based Augmentation (IHTA), which synthesizes the methodologies of hierarchical generalization of demographic terms and threshold-based augmentation techniques. This synthesis aims to improve the accuracy and fairness of commonsense polarization classifiers.

The IHTA framework begins by employing augmentation techniques on sentences where demographic terms exceed predetermined thresholds, followed by hierarchical generalization.

This combined method is designed to scrutinize and compare the performance of classifiers when the methodologies are employed either in isolation or in their integrated form.

### IV. Experiment

#### A. Analysis of Hierarchical Generalization

In this section, we examine the effects of applying hierarchical generalization and threshold-based augmentation on the performance of our commonsense bias classifier.

#### 1) BASELINES VS GENERALIZATION

Drawing upon the findings, which investigated the impact of demographic terms, it was determined that the decline in performance was attributed to demographic factors. Consequently, this study employed a hierarchical generalization approach to address the issue of performance degradation resulting from the influence of demographic terms. The performance of the classifier was evaluated by comparing it before and after implementing hierarchical generalization. The assessment examined the outcomes in terms of accuracy, precision, recall, and F1-score. The results, as shown in Table 2, demonstrate that hierarchical generalization attained higher accuracy values in comparison to the baseline, exhibiting an improvement of 2.73% for the total dataset. Likewise, for the Unique dataset, hierarchical generalization outperformed the baseline by 1.26%. The unique dataset, comprising 3,227 sentences, was evaluated by three workers, yielding a total of 10,581 labeled sentences. The accuracy increase over the baseline is 2.33%. During the transition from the baseline to generalization1, there was an average increase of 0.23% for the total and an average increase of 1.57% for the unique. Furthermore, when progressing from generalization1 to generalization2, there was an average increase of 5.23% for the total and an average increase of 0.95% for the unique. The F1 values exhibit similar patterns. The increase in F1 score over the baseline is 2.21%. During the transition from the baseline to generalization1, an average increase of 0.41% for the total F1 score and an average increase of 2.00% for the unique F1 score were observed. Furthermore, in the progression from generalization1 to generalization2, there was an average increase of 5.41% for the total F1 score and an average increase of 1.50% for the unique F1 score. Overall, the performance seems to improve as the generalization increases through the hierarchy, at least up to the level where the general category can be identified. In Fig.8., 'Correctly classified' indicates agreement between human annotators and the BERT model's classifications, whereas 'Misclassified' denotes instances where the BERT model's classifications differ from human annotations. For example, as depicted in Fig. 8., the baseline model misclassified the sentence containing the demographic term "boy" whereas both hierarchy1 and hierarchy2 achieved accurate classification for that sentence. Similarly, the baseline and hierarchy1 struggled with classifying the sentence containing the demographic term "jihad" whereas hierarchy2 correctly classified it.

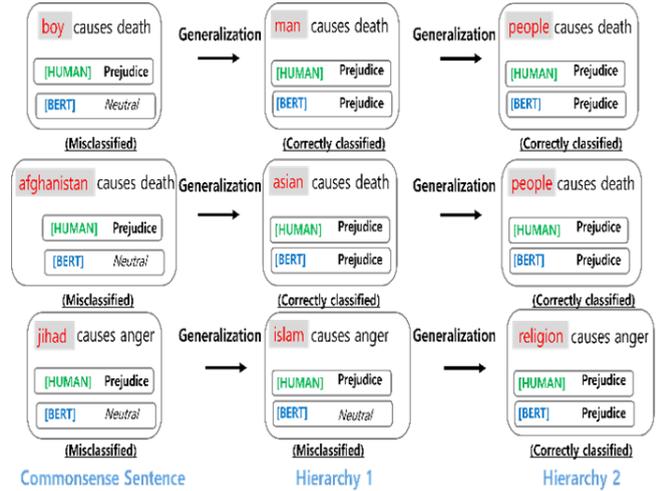

**FIGURE 8.** Effect of Hierarchical Generalization on Performance in Commonsense Sentence Classification.

As the generalization hierarchy progresses, the demographic terms become more generalized, resulting in improved performance of the polarity classifier. However, when examining the generalization outcomes across religion, origin, profession, and gender classes, there was limited increase in the accuracy with an average increase of 0.28% observed across the classes in the dataset.

### 2) RELATION BETWEEN DATASET AND ACCURACY

In the previous experimental results, it was observed that increasing the generality led to an improvement in the performance of the polarization classifier for large dataset sizes, both with the total or unique datasets. However, for the categories with a smaller number of datasets, the improvement was found to be limited. We hypothesize that the observed differences in performance can be explained by the correlation between the size of the datasets and performance. Fig. 9. illustrates the correlation between the number of datasets and accuracy. The result is a statistically significant positive correlation ($r = 0.35$, $p < 0.001$). Based on this analysis, we propose a threshold-based data augmentation when working

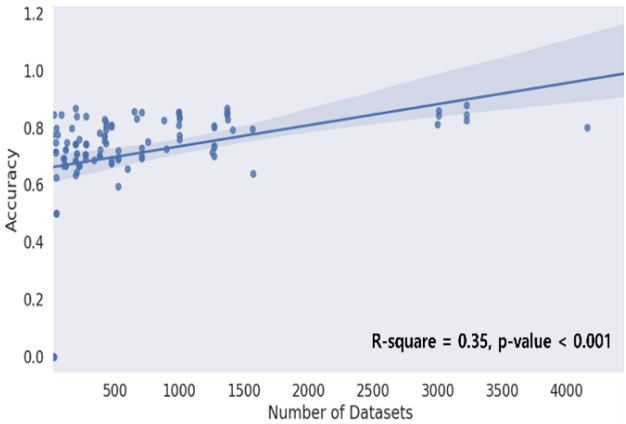

**FIGURE 9.** Correlation between Number of Datasets and Accuracy in Polarization Classifier.

with a small number of commonsense datasets.

### 3) THRESHOLD-BASED DEMOGRAPHIC TERM MEASURING

The results obtained from measuring polarization using the threshold-based technique are presented in Table 3, with all decimal values rounded to two decimal places. In the 'Origin' category, the highest polarization was observed for Africa (-0.14), followed by America (-0.12), Asia (-0.11), and Europe (-0.09). The average polarity of the 'Origin' category was -0.11, suggesting an overall negative perception.

In the 'Religion' category, the highest polarization was observed for Christianity (0.08), followed by Hinduism (-0.13) and Islam (-0.08), indicating a preference for Christianity and negative perceptions towards Hinduism and Islam. The average polarity of the 'Religion' category was -0.04, indicating an overall negative perception. In the 'Profession' category, the highest polarization was observed for Industrial & Manufacturing (-0.05), followed by Science & Technology (-0.04), Art & Entertainment (0.03), and Healthcare & Medicine (0.04). There was a preference for Healthcare & Medicine and a perception of negative towards Industrial & Manufacturing. The average polarity of the 'Profession' category was -0.02. These findings reveal diverse levels of polarization and underscore the importance of mitigating biases and promoting equity in commonsense data and language models.

**TABLE 3.** Measurement of Polarization Using Threshold-based Technique.

| | Class(polar) |
|---|---|
| Origin | Europe (-0.09), Asia (-0.11), Americas (-0.12), Africa (-0.14) |
| Religion | Christian (0.08), Hindu (-0.13), Muslim (-0.08) |
| Gender | Male (-0.05), Female (-0.11) |
| Profession | Healthcare & Medicine (0.04), Art & Entertainment (0.03), Science & Technology (-0.04), Industrial & Manufacturing (-0.05) |

### 4) DATA AUGMENTATION RESULTS

We employed a novel threshold-based augmentation technique that quantifies the polarization of demographic terms. This was achieved by evaluating the differences in the model's predictions when these terms were masked versus unmasked. Additionally, we applied the augmentation to commonsense sentences with polar values by substituting their tuple relations with synonyms generated by ChatGPT. An experiment focusing on the religion and origin classes, which had a relatively small number of datasets, was conducted. This analysis compared the performance of the classifiers with threshold-based augmentation to that with standard augmentation, focusing on metrics such as accuracy.

Fig. 10. illustrates the performance trends within the religion class. The threshold-based augmentation demonstrated an average accuracy of 0.8555, outperforming the standard augmentation, which had an average accuracy of 0.8418. Furthermore, the threshold-based augmentation exhibited a significantly higher correlation coefficient of 0.91 between the number of datasets and accuracy metrics, compared to the correlation coefficient of 0.79 achieved by the standard augmentation technique ($p < 0.05$).

The IHTA (Integrated Hierarchical Generalization and Threshold-based Augmentation) method's results, as illustrated in Fig. 10, showed that the average accuracy for the combination of threshold-based augmentation and Generalization1 was 0.9534, and for the combination with Generalization2, it was 0.9469. As the dataset size increases, the IHTA approach with Generalization 2 consistently achieves higher accuracy compared to Generalization 1. Meanwhile, compared to the correlation coefficient of 0.6246 achieved by the combination of threshold-based augmentation and Generalization1, the combination with Generalization2 exhibited a significantly higher correlation coefficient, 0.6591, between the number of datasets and accuracy metrics ($p < 0.05$).

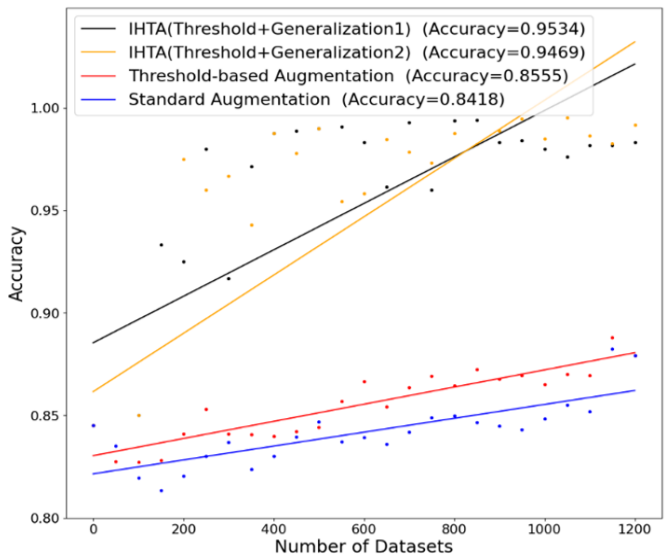

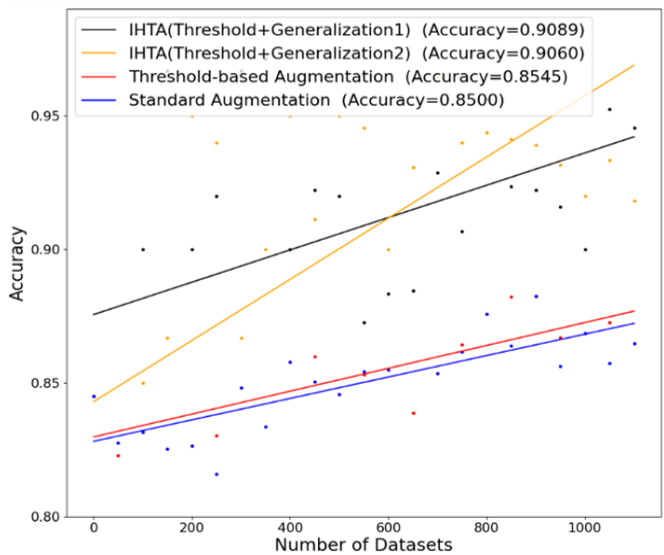

**FIGURE10.** Correlation between Number of Datasets and Accuracy in Polarization Classifier.

Within the religion class, the combined methods presented a higher average accuracy than the threshold-based method alone, showing an improvement of 9.90%, and a higher average accuracy than the standard augmentation method alone, with an improvement of 11.27%.

Additionally, within the gender class, the threshold-based augmentation technique exhibited an average accuracy of 0.8545, exceeding the average accuracy of 0.8500 achieved by the standard augmentation technique, as illustrated in Fig. 10. Moreover, the threshold-based augmentation displayed a significantly higher correlation coefficient, 0.8213, between the number of datasets and accuracy metrics, compared to the correlation coefficient of 0.8014 achieved by the standard augmentation technique ($p < 0.05$).

Within the gender class, the IHTA method's results indicated that the average accuracy for the combination of threshold-based augmentation and Generalization1 was 0.9089, and for the combination with Generalization2, it was 0.9060. Furthermore, this trend of increased accuracy with larger datasets reinforces the effectiveness of Generalization 2 over Generalization 1 in the IHTA approach. Meanwhile, the combination of threshold-based augmentation and Generalization2 exhibited a significantly higher correlation coefficient, 0.5196, between the number of datasets and accuracy metrics, compared to the correlation coefficient of 0.3693 achieved by the combination with Generalization1 ($p < 0.05$).

Within the gender class, the combined methods presented higher average accuracy than threshold-based method alone, with an improvement of 7.74% and higher average accuracy than Standard Augmentation method alone, with an improvement of 8.65%.

These results demonstrate that the performance enhancement achieved through the combined techniques was more pronounced compared to the effects of standalone threshold-based augmentation. There is a performance increase compared to the use of threshold-based augmentation or standard augmentation alone.

## V. Conclusion

In this study, we proposed two methods to enhance commonsense bias classifiers by mitigating the impact of demographic terms. The first method, Hierarchy Generalization, involved replacing specific demographic terms with more general terms based on a predefined hierarchy ontology, resulting in a 2.33% performance increase over the baseline. The second method, threshold-based data augmentation, measured the polarization of demographic terms and applied augmentation to commonsense sentences. This method outperformed standard data augmentation by 0.96%. The two methods we proposed, while resulting in relatively modest accuracy improvements of 2.33% and 0.96% respectively, are of significant importance in the context of mitigating bias in NLP models.

The combined techniques (IHTA) showed higher average accuracy than threshold-based method alone, with an improvement of 8.82% and higher average accuracy than standard augmentation method alone, with an improvement of 9.96%.

These improvements contribute to making the polarization classifier more accurate and equitable, which is crucial for reducing the impact of demographic terms on model predictions and ensuring fairness in NLP. Overall, our study suggests that hierarchical generalization and threshold-based augmentation can potentially mitigate bias issues and improve NLP model performance.

## VI. Limitation and Future work

Despite the promising results and contributions of this study, there are several limitations that should be acknowledged. Firstly, the study primarily focuses on demographic biases related to gender, religion, origin, and profession, while other forms of bias, such as age, disability, and sexual orientation, have not been explicitly addressed. Future research should explore extending the proposed approach to encompass a wider range of demographic domains and intersectional biases. Further investigation is needed to assess the generalizability of the results to different datasets and models, ensuring the reliability and effectiveness of the approach across diverse applications.

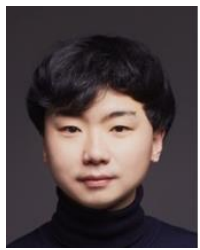

**JINKYU LEE** received his B.S. and M.S. degrees in Forest Engineering from Kookmin University, South Kore a, in 2018. He is currently pursuing a Ph.D. degree at the Graduate School of Artificial Intelligence. His current research interests include natural language processing and AI bias.

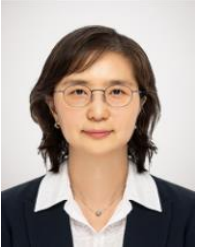

**JIHIE KIM** (received the B.S. degree in computer science and statistics and the M.S. degree in computer science and statistics from Seoul National University in1988 and 1990, respectively, and the Ph.D. degree in computer science from the University of Southern California, in 1996. She is currently a Professor of the Division of AI Software Convergence, Dongguk University, Seoul, South Korea. Her research interests include machine learning, NLP, and knowledge-based reasoning.